\documentclass{article}
\usepackage{spconf,amsmath,graphicx}
\usepackage{rotating}
\usepackage[moderate]{savetrees}

\usepackage[compact]{titlesec}
\titlespacing{\subsubsection}{0pt}{0.6ex}{0ex}

\newcommand{\specialcell}[2][c]{\begin{tabular}[#1]{@{}c@{}}#2\end{tabular}}

\title{Feature Imitating Networks}

\name{Sari Saba-Sadiya$^{\star}$ \qquad Tuka Alhanai$^{\dagger}$ \qquad Mohammad M Ghassemi$^{\star}$}
  
  \address{$^{\star}$ Department of Computer Science, Michigan State University,
               East Lansing, MI\\
      $^{\dagger}$ Department of Computer Engineering, 
               New York University, Abu Dhabi, UAE}
      
\begin{document}
%
\maketitle
\begin{abstract}
In this paper, we introduce a novel approach to neural learning: the Feature-Imitating-Network (FIN). A FIN is a neural network with weights that are initialized to reliably approximate one or more closed-form statistical features, such as Shannon's entropy. In this paper, we demonstrate that FINs (and FIN ensembles) provide best-in-class performance for a variety of downstream signal processing and inference tasks, while using less data and requiring less fine-tuning compared to other networks of similar (or even greater) representational power. We conclude that FINs can help bridge the gap between domain experts and machine learning practitioners by enabling researchers to  harness  insights  from  feature-engineering to enhance the performance of contemporary representation learning approaches.

\end{abstract}

\section{Introduction}
The successful application of deep learning to new problem domains has three conditions: (1) access to large data sets, (2) access to sufficient computing resources for hyper-parameter optimization and (3) modest expectations about model interpretability. Deep learning models require large data-sets to learn representations that generalize on future unseen data. Additionally, extensive exploration of the model topological space is often necessary to identify a network architecture with sufficient representational power for a given task. Lastly, despite ample recent work on interpretability of deep learning models, the community remains without normative standard for how deep networks should be interpreted; this is problematic for many problem domains (e.g. healthcare) where the importance of interpretability may supersede performance \cite{Roberts_2021}. 

\subsection{Contributions}
In this paper we introduce Feature-Imitating-Networks (FINs): a FIN is a neural networks with weights that are initialized to approximate one or more closed-form statistical features. In this paper, we will demonstrate how this property of FINs improves their interpretability while also reducing data and hyper-parameter tuning requirements compared to other networks with similar or greater representational power. More specifically, we demonstrate how, when combined with a careful application of transfer-learning, and by taking into account expert knowledge, FINs can be used to quickly build and deploy robust and better performing models using less training epochs. Our validation of FINs focused on tasks involving biomedical signals; the data-sets in this domain are often smaller, and therefore stand to benefit the most from the introduction of our framework.

\subsection{Paper Organization}
The remainder of the paper is organized as follows: First we review relevant literature regarding transfer learning. The \textit{Related Work} section is followed by the \textit{Methodology} section where we discuss how to build and design different FINs. The \textit{Experiments} section contains three experiments - including a brief discussion of the data and results for each. Finally, the \textit{Discussion} section examines all the results in aggregate and discusses how our framework might be expanded. 

\section{Related Work} \label{sec:Related}
Transfer learning is the application of a pre-trained model to tasks it was not originally intended to perform \cite{Ng_TL}. Transfer learning enabled researchers to make significant progress on various tasks in Machine Vision\cite{Hon_2017}, Speech \cite{Lech_2020}, and Natural Language Processing \cite{ULMFiT}.

Most applications of transfer learning are \textit{within-domain}; these involve fine-tuning a pre-trained model for new tasks. For instance, AlexNet \cite{AlexNet}, VGG \cite{VGG16}, and ResNet \cite{resNet50} are computer vision models that were trained to classify the ImageNet data-set. The features learned by these models (in later layers), and their more fundamental image components (in earlier layers) can be re-purposed to solve other tasks using only a fraction of the training data required by the original models. 

Models that are trained on large heterogeneous data-sets are good candidates for "transfer". But for biomedical signal processing problems, there isn't a sufficiently large data-set to train such a model. Indeed, the largest publicly available biomedical signal archives contain only a few thousand subjects, which is too small by most data standards in other domains \cite{ghassemi2018you}. Consequently, transfer learning for biomedical signals is often performed \textit{across-domains}. For instance, computer vision models such as VGG have been adapted to emotion recognition from speech \cite{Lech_2020}, motor-imagery classification \cite{Xu_2019} and mental task classification \cite{Opalka_2018}, but these \textit{cross-domain} transfers are not as effective as those performed \textit{within-domain}.

The performance of transfer learning is proportional to the proximity of the domains across which knowledge is being transferred. Feature imitating networks were designed to address this limitation of current transfer learning paradigms. - they provide the power and flexibility of transfer learning without the  ``Big data'' and heavy computational requirements.

\section{Methodology} \label{sec:Meth}
A FIN is a neural network with weights that are initialized to approximate one or more closed-form statistical features. In this paper, we train FINs that approximate five commonly used features in biomedical signal processing: Shannon's Entropy, kurtosis, skewness, fundamental frequency, Mel-frequency cepstral coefficients (mfcc), and regularity \cite{saba2020unsupervised}. We evaluate the utility of the FINs on three biomedical signal processing experiments, which we describe in Section 4, below. The pre-trained FINs, and code to reproduce all experimental results are available online \footnote{https://github.com/tobe/released/following/acceptance}. 

\begin{figure}[t]
\centering
\includegraphics[width=0.83\columnwidth]{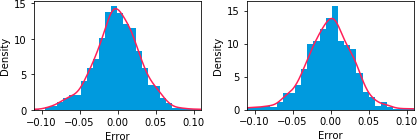} 
\caption{Density plots for the errors for the entropy (left) and regularity (right) FIN reconstructions. The feature values were scaled and normalized, making the biggest possible error 1, as can be observed the FINs faithfully recreate the closed form equations.}
\label{fig:meth}
\end{figure}

\paragraph*{Network Construction}
For each feature, we used a simple gradient descent optimizer with mean square error (MSE) loss to train a simple dense network to approximate that feature on synthetic signals. The topological space explored for all the FINs was between $2$ to $10$ layers with the number of parameters in the $3-15$ million range. All best performing FINs used simple $relu$ and $tanh$ activation functions. See Fingure \ref{fig:meth} for density functions for the errors of the FINs reconstruction.

\paragraph*{Input} 
The input data for the FINs consisted of synthetic signals generated randomly (zero mean, unit variance) and converted to the time-frequency domain using the wavelett transform \cite{Lech_2020,Xu_2019}. 

\paragraph*{Outcome} 
The outcome data for the FINs consisted of closed-form feature values calculated on the synthetic signals using \textit{SciPy} and \textit{EEGExtract} packages \cite{saba2020unsupervised}. 

\paragraph*{Transfer} 
When applying the models to new classification tasks, (i.e. Section \ref{sec:Exp}) the very last layer was discarded in favor of a randomly initialized softmax layer with dimensions suitable to the task.

\paragraph*{Baseline model}
The baseline in each experiment was the best performing neural network with similar (or greater) representational power to the corresponding FIN, trained using the same training data and schedule, but with weights that were randomly initialized; In total one hundred different topologies were explored for each baseline. The baseline with the best average performance on the validation data was retained for comparison against the FINs. 

\paragraph*{Training}
All models (both FIN and Baseline) were trained using early stopping and a simple gradient descent optimizer. Non topological hyper-parameters such as learning rate and momentum had minor effects in comparison and therefore will be omitted from future discussions. Training was conducted using \textit{CUDA} on a \textit{Tesla K80} GPU with $25$GB of RAM. 

\vspace*{-2mm}

\section{Experiments} \label{sec:Exp}
To evaluate our framework we ran three experiments on three different biomedical data-sets and tasks. The first experiment was an Electrocardiography (ECG) classification task; our goal was only to demonstrate that our FINs framework can successfully improve performance on small low-accuracy data-sets. The second experiment was an Electroencephalography (EEG) artifact detection task; our goal was to demonstrate the modular nature of FINs, and the potential benefits of using FIN ensembles. The third experiment was a drowsiness detection task using EEG; our goal was to compare both the performance and speed of FINs against state-of-the-art transfer-learning techniques under conditions of varying data scarcity. 

For all three experiments, the data was regularized and transformed to the time-frequency domain as discussed in the Methods section. The data was partitioned into training, validation, and testing sets. In the first two experiments this was achieved by randomly partitioning the data $15\%-85\%$ for testing and training respectively, before repeating the same split for the training data to extract a validation subset. This was repeated for a 50-fold cross-validation. In the third experiment, where subject data is balanced, the partitioning was achieved by iteratively leaving two of the twelve subjects out for validation and testing. To compare training time we used similar instances of nodes with \textit{Tesla K80} GPUs and $25$GB of RAM, all training times are reported in seconds.

\subsection{Experiment I}
In this experiment, we explored the potential of FINs for the detection of artifact ridden ECG signals \cite{nemcova2021brno,PhysioNetComp}.

\paragraph*{Data and Prepossessing} We used data made available by The \textit{Brno University of Technology ECG Quality Database} \cite{nemcova2021brno}. ECG segments of variable lengths from 18 subjects were classified by experts into three categories according to signal quality. After standardizing the lengths we ended up with $2544$ trials. Preprocessed data will be made available.

\paragraph*{Models} We hypothesized that a FIN trained to imitate Kurtosis might be useful in the context of this task \cite{Zhao_2018}. The Kurtosis FIN was adapted for our classification task by replacing the very last layer with a softmax classification layer. In addition to the baseline neural network, we also compare against several non deep learning classification algorithms.

\paragraph*{Results} As can be seen in Table \ref{tab:exp1}, The Kurtosis FIN consistently outperformed the baseline models. Moreover, the standard deviation in the performance of the FINs was an order of magnitude smaller than the deep network baseline models. A Levene's test indeed indicates a statistically significant ($p<.05$) difference in variance between the performance of the two methods throughout the iterations. This highlights the fact that our framework helps with the robustness of the models.

\begin{table}[t]
\centering
\small
\begin{tabular}{|c|c|c|c|c|}
    \hline
    Model & Baseline & SVM & kNN & \specialcell{Fine-tuned  \\ FIN} \\
    \hline\hline
    \specialcell{Mean \\ ($\pm$ std)}  & \specialcell{$.443$  \\ $(\pm 0.174)$} & \specialcell{$.5233$  \\ $(\pm 0.016)$} & \specialcell{$.525$  \\ $(\pm 0.018)$} & \specialcell{$.543$ \\ $(\pm .0245)$} \\ 
    \hline
\end{tabular}
\caption{Mean and standard deviation of the accuracy for the Experiment I classification task. As demonstrated the FIN based network out performs both randomly initialized Neural Networks and classical statistical approaches.}
\label{tab:exp1}
\end{table}

\begin{table*}[t]
\centering
\small
\begin{tabular}{|c|c|c|c|c|c|c|c|c|c|}
    \hline
    FIN & Regularity & \specialcell{Fundamental \\ Frequency} & \specialcell{Entropy+\\Regularity} & \specialcell{Kurtosis+\\ Regularity} & \specialcell{Baseline} & \specialcell{MFCC} & \specialcell{Entropy} & \specialcell{Kurtosis} & \specialcell{Entropy+Kurtosis\\ +Regularity}\\
    \hline\hline
    \specialcell{Mean \\ ($\pm$ std)} & \specialcell{$.6527$  \\ $(\pm .1066)$} & \specialcell{$.6825$  \\ $(\pm .0591)$}  & \specialcell{$.6991$  \\ $(\pm .0662)$}  & \specialcell{$.7134$  \\ $(\pm .0807)$} & \specialcell{$.7142$  \\ $(\pm .0587)$} & \specialcell{$.7167$  \\ $(\pm .01411)$} & \specialcell{$.7195$  \\ $(\pm .03783)$} & \specialcell{$.0.7214$  \\ $(\pm .02397)$} &\specialcell{$.0.724$  \\ $(\pm .0451)$} \\ 
    \hline
\end{tabular}
\caption{Mean and std of the accuracy for experiment 2. Corrected one-tailed t-tests demonstrated that models imitating features known to be useful for EEG artifact detection (last three rows) significantly out-performed models imitating ill-suited features (first two rows).}
\label{tab:exp2}
\end{table*}

\subsection{Experiment II}
In this experiment, we investigate how different FINs can be used in conjunction to build complex networks suited for EEG artifact detection. Moreover, we demonstrate how theoretical knowledge regarding the features and their relevancy to the task is helpful when using the FINs Framework.

\paragraph*{Data and Prepossessing} The data used in this experiment is from an EEG artifact detection data-set \cite{saba2020unsupervised}. The data contains EEG segments from a $1kHz$ recording made using $32$ electrodes during a passive viewing task. Each segment is a second long and was labeled as artifact ridden or clean by expert annotators. We re-sampled the data at $500Hz$ and converted the EEG setup to the international $10-20$ system that contains only $19$ electrodes. 

\paragraph*{Models} 
We evaluated individual FINs and FIN ensembles trained to imitate Kurtosis, Shannon's Entropy, Regularity, Fundamental Frequency, MFCCs, and ensemble combinations thereof. We expect some of these FINs to outperform others based on the task-relevance of the feature being imitated. For instance, the fundamental frequency of the signal, defined as the lowest periodic frequency of the waveform should be irrelevant, while the Kurtosis is highly relevant to the task \cite{Delorme_2001,Javidi_2011}. Similarly, we expect `Complexity Features' such as The Cepstrum Coefficients and Shannon's entropy to outperform clinically grounded `Continuity Features'  such as the fundamental frequency or EEG regularity (burst suppression) \cite{saba2020unsupervised}. Following these theoretical considerations we hypothesize that the MFCC, Entropy, and Kurtosis FINS, as well as a combination of these FINS will outperform the Regularity FINs. As we have multiple hypotheses used the appropriate Bonferroni correction for multiple comparisons. To have enough statistical power after the correction we increased sample size by repeating each experiment $50$ times.

Each FIN was applied on each electrode signal in parallel, the outputs were then concatenated and passed forward to a binary softmax classification layer. We compared the FINs against the best performing baseline dense neural network.

\paragraph*{Results} As summarized in Table \ref{tab:exp2}, our experiment demonstrates how (when appropriately selected) FIN ensembles may be used in combination to further enhance task performance. We note here that deliberate consideration when combining FINs can improve task performance, while keeping the size of the ensemble small. 

A corrected one tailed t-test showed that after correction the Kurtosis, Entropy, and Ensamble Network (last three rows in the table) performed significantly better than the Fundamental Frequency or Regularity FINs.

\subsection{Experiment III}
In this experiment, we compare the performance of FINs against state-of-the-art approaches for a Fatigue and drowsiness detection from EEG task on a recently published data-set.

\paragraph*{Data and Prepossessing} We used data made available by \cite{Jianliang_2017}. This paper identified multiple subsets of electrodes as especially predictive. We tested separately on every subset. The data was then partitioned for six fold intra-subject cross validation. In other words, at each iteration ten out of twelve available subjects were used for training, one was used for validation, and the testing accuracy was calculated on the remaining subject. Finally, each cross-validation step was repeated $5$ times. If only a fraction of the training data was being used different trials were picked at each of these iterations. 

\paragraph*{Models} 
Prior work indicates that entropy is a useful feature for the prediction of drowsiness and fatigue from EEG data \cite{Jianliang_2017}. Thus, we compared a FIN trained to imitate Shannon's Entropy against the baseline models (described in the methods), as well as a fine tuned VGG network pre-tarined on the ImageNet data-set \cite{VGG16}. The VGG model was similar to the $19$ layer convolutional neural network introduced in \cite{VGG16} sans the last $3$ dense layers and with the addition of a final softmax classification layer. This is the standard way in which the VGG model is used in biomedical tasks \cite{Opalka_2018,Xu_2019}.

Additionally, to test the performance of our pre-trained FIN when only very limited data is available we ran the same models but with varying fractions of the data being made available during training. 

\begin{figure}[t]
\centering
\includegraphics[width=0.85\columnwidth]{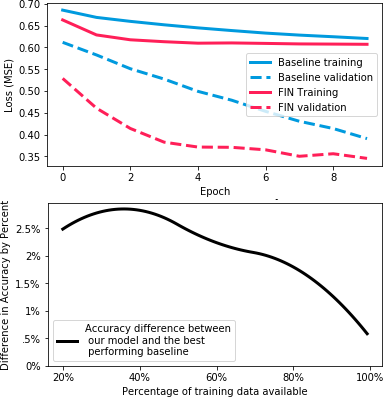} 
\caption{Experiment 3 results. (top): Training and Validation loss for the baseline and pre-tuned entropy FIN. (Bottom): Difference in accuracy between our pre-trained FIN and the best performing baseline as a function of the percentage of data available.}
\label{fig:exp3}
\end{figure}

\paragraph*{Results} The pre-tuned Shannon entropy FIN outperformed the baseline and reinitialized FIN in each of the four subsets and at over $83\%$ of the iterations of the cross validation. Additionally, as can be seen in Figure \ref{fig:exp3} (top), the pre-tuned FIN had lower loss at every epoch compared to the baseline. It is important to note that the FIN also beat the performance of classical classifiers with different entropy measures that was reported in the literature \cite{Jianliang_2017}.

The performance improvements were even greater under limited data availability conditions As small training data sets are known to increase performance noise, we also repeated this process $10$ times and reported the average accuracy; the difference in the average accuracy between our method and the baseline for each data percentage is plotted in figure \ref{fig:exp3} (bottom). 

The VGG transfer learning network under-performed both the FINs and other baseline models and was particularly sensitive to small data sizes. When only a fraction of the data was available, VGG performed at close to chance.

\begin{table}[t]
\centering
\small
\resizebox{.99\columnwidth}{!}{
\begin{tabular}{|c|c|c|c|}
    \hline
    \specialcell{Mean ($\pm$ std)\\ Training Seconds} & Baseline & VGG & FIN \\
    \hline\hline
    $20\%$ of data & \specialcell{$.746 (\pm 0.019)$ \\ $37.2 (\pm 8.9)$} & \specialcell{$0.615 (\pm 0.122)$ \\ $21911 (\pm 520)$} & \specialcell{$\mathbf{.771 (\pm 0.016)}$ \\ $50.8 (\pm 5.7)$} \\
    \hline
    $40\%$ of data & \specialcell{$.895 (\pm 0.04)$ \\ $37.2 (\pm 0.2)$} & \specialcell{$.644
 (\pm 0.18)$ \\ $10881 (\pm 1441)$} & \specialcell{$\mathbf{.922 (\pm 0.013)}$ \\ $106.0 (\pm 23.3)$} \\
    \hline
    $60\%$ of data & \specialcell{$ .939 (\pm .07)$ \\ $ 189.8 (\pm 79.9)$} & \specialcell{$ .69 (\pm .197)$ \\ $ 18434 (\pm 1058)$} & \specialcell{$\mathbf{ .962 (\pm .006)}$ \\ $ 322.4 (\pm 201.2)$} \\
    \hline
    $80\%$ of data & \specialcell{$ .983 (\pm 0.0203)$ \\ $ 574.8 (\pm 132.3)$} & \specialcell{$ .802 (\pm .103)$ \\ $ 17141 (\pm 1838)$} & \specialcell{$\mathbf{ .996 (\pm .0013)}$ \\ $ 441.6 (\pm 92.2)$} \\
    \hline
    $100\%$ of data & \specialcell{$ .993 (\pm 0.022)$ \\ $ 645.7 (\pm 187.1)$} & \specialcell{$ .846 (\pm .088)$ \\ $ 19267.3 (\pm 2014)$} & \specialcell{$\mathbf{ .998 (\pm 0.001)}$ \\ $ 552.6 (\pm 129.7)$}\\
    \hline
\end{tabular}}
\caption{Experiment 3 results. The models were trained using varying subsets of the data. We report the mean and standard deviation for both accuracy and training duration on a node with a \textit{Tesla K80} GPU and $25$GB of RAM running \textit{CUDA}}
\label{tab:exp3}
\end{table}

\section{Discussion} \label{sec:Disc}
The feature imitating networks framework proposed in this paper is an innovative way to use transfer learning. Traditional transfer learning requires large, slow to train, black-box, networks such as \textit{VGG} and \textit{AlexNet} tuned on hundreds of thousands of labeled data. In contrast, FINs require no human labeling, are small and fast to train, and can be combined to create ensemble FIN networks in accordance with insights from the literature surrounding the task being performed. Therefore, our network facilitates the integration of domain specific knowledge into modern data driven machine learning practices. Beyond these considerations there are several practical benefits to using our framework:

\begin{itemize}
\item \textbf{Robustness:}
Our experiments indicated that FINs are more robust than other networks and techniques with similar representational power. This is evident in statistically significant differences in variations in accuracy when performing leave subject out and cross validation. Deep learning in general is sensitive to weight initialization randomness and data idiosyncrasies. Transfer learning of weights tuned to calculate task relevant features seems to guarantee we start at a 'neighborhood` of a good solution. Moreover, FINs expedite the hyper-parameter optimization step which remains resource-consuming despite recent research \cite{Wu_2019,Miikkulainen_2019}.    

\item \textbf{Performance:}
Data scarcity still plagues many domains. In the case of biomedical research data collection is especially costly and can prohibit researchers from applying deep learning to their tasks altogether \cite{Sadiya_NER21}. Our experiments indicate that FINs are useful especially when only limited data is available. The intuition behind this is straightforward; pre-trained weights already extract useful task-relevant information, resulting in a better performance and lower loss when from the very first epochs of the training procedure, as can be seen in Figure \ref{fig:exp3}.

\item \textbf{Flexibility:}
By tuning on task-specific data sets our framework also out performs methods that pass the calculated features as input to the classifiers. This is not surprising as our FINs are allowed to tune the extracted features to better suit the task (for instance by focusing on specific parts of the signal). Additionally, the modular nature of the FINs lends itself to easily building and testing ensembles networks.      

\item \textbf{Speed:}
\textit{VGG} and \textit{AlexNet} are powerful networks that have been successfully applied in various domains. However, these architectures are extremely large. The \textit{VGG} based network consists of at least $17$ layers and contains over $20$ million parameters. In contrast, FINs are simple shallow networks consisting of up to $4$ layers and a quarter of that numbers of parameters. The shallowness of the models in particular guarantees that even when using an ensemble of FINs gradient descent calculations, and therefore training and inference times, remain simple and fast. This can be observed in the results of the third experiment presented in Table \ref{tab:exp3}, training the $VGG$ network was in some cases over $60$ times slower than the FINs network training despite lower performance.

\subsection{Future Directions}
Designing the dense implementation of the FINs can be streamlined by considering the closed form equation of the signal and training layers to imitate each operator separately. For instance, the mathematical expression for Shannon's entropy requires discretization of the signal to create a histogram before averaging each bin. Partitioning the operations allows us to reuse pre-trained operation-specific layers to quickly construct FINs that are then fine tuned to mimic specific features.

\section{Conclusion} \label{sec:Con}

In recent years, some have critiqued the current state of the machine learning community. These critiques often focus on disregard of traditional techniques in favor of data driven approaches \cite{manningDeepLearning} and the different ways deep learning have struggled to live up to it's promise, especially when it comes to real world applications \cite{Roberts_2021}. In this paper, we presented \textit{Feature-Imitating-Networks}, a variation over traditional transfer learning that uses networks trained to imitate simple closed form statistical features, that we believe elevates these concerns. We demonstrated that our framework is superior in both the speed and accuracy to deep and transfer learning techniques with similar (or greater) representational ability. Especially when only very limited data is available. The experiments were conducted on a variety of tasks and domains. Future work will extend this initial exploratory work. An extensive library of Feature Imitating Network bench-marked on many data-sets and other signal processing domains might be of particular use and interest to the research community.
\end{itemize}
\newpage

\bibliographystyle{IEEEbib}
\bibliography{refs}

\end{document}